\documentstyle[eaclap]{article}

\author{
Hinrich Sch\"{u}tze \\
CSLI, Ventura Hall \\
Stanford, CA 94305-4115 , USA \\
email: {\tt schuetze@csli.stanford.edu}\\
 URL: {\tt ftp://csli.stanford.edu/pub/prosit/DisPosTag.ps}
}

\title{Distributional Part-of-Speech Tagging}

\begin{document}

\maketitle
\vspace{-0.5in}
\begin{abstract}
This paper presents an algorithm for tagging words whose
part-of-speech properties are unknown. Unlike previous work,
the algorithm categorizes {\em word tokens in context\/}
instead of {\em word types\/}.  The algorithm is
evaluated on the Brown Corpus.
\end{abstract}

\bibliographystyle{acl}

\section{Introduction}
Since online text becomes available in ever increasing volumes and an
ever increasing number of languages,
there is a growing need for robust processing
techniques that can analyze text without expensive and time-consuming
adaptation to new domains and genres. This need motivates research on
fully automatic text processing that may rely on general principles of
linguistics and computation, but does not depend on knowledge about
individual words.

In this paper, we describe an experiment on fully automatic derivation
of the knowledge necessary for part-of-speech tagging. Part-of-speech
tagging is of interest for a number of applications, for example
access to text data bases \cite{kup93}, robust parsing \cite{abn91},
and general parsing \cite{dem90,cha94}.  The goal is to find an
unsupervised method for tagging that relies on general distributional
properties of text, properties that are invariant across languages and
sublanguages. While the proposed algorithm is not successful for all
grammatical categories, it does show that fully automatic tagging is
possible when demands on accuracy are modest.

The following sections discuss related work, describe the learning
procedure and evaluate it on the Brown Corpus
\cite{fk82}.

\section{Related Work}
The simplest part-of-speech taggers are bigram or trigram models
\cite{chu89,cha93}. They require a relatively large
tagged training text. Transformation-based tagging as
introduced by \newcite{bri93} also requires a hand-tagged text for
training.  No pretagged text is necessary for Hidden Markov Models
\cite{jel85,cutting91,kup92}. Still, a lexicon is needed that
specifies the possible parts of speech for every word. \newcite{bm93}
have shown that the effort necessary to construct the part-of-speech
lexicon can be considerably reduced by combining learning procedures
and a partial part-of-speech categorization elicited from an
informant.

The present paper is concerned with tagging languages and sublanguages for
which
no a priori knowledge about grammatical categories
is available, a situation that occurs often in
practice \cite{bm93}.

Several researchers have worked on learning grammatical
properties of words. \newcite{elm90} trains a connectionist net to
predict words, a process that generates internal representations that
reflect grammatical category.
\newcite{bmms90} try to infer grammatical category from bigram statistics.
\newcite{fc92} and \newcite{fin93} use
vector models in which words are clustered according to the similarity
of their close neighbors in a corpus. \newcite{kn93} present a
probabilistic model for entropy maximization that also relies on the
immediate neighbors of words in a corpus. \newcite{bib93} applies
factor analysis to collocations of two target words (``certain'' and
``right'') with their immediate neighbors.

\begin{table*}[tbp]
\small
\centering
\begin{tabular}{l|l|l}
{\bf word} & {\bf side} & {\bf nearest neighbors} \\\hline\hline
   onto &left &into  toward  away  off
together    against beside around down\\
  onto &right &reduce among regarding against
towards plus  toward using
unlike\\\hline
 seemed &left &appeared might would remained
had became could must   should\\
 seemed    &right &seem   seems   wanted   want
going    meant   tried   expect   likely
\end{tabular}
\caption{\label{nearest} Words with most similar left and right
neighbors for ``onto'' and ``seemed''.}
\end{table*}

What these approaches have in common is that they classify {\em
words\/} instead of individual occurrences.
Given the widespread
part-of-speech ambiguity of words this is
problematic.\footnote{Although
\newcite{bib93} classifies collocations, these can also be ambiguous.
For example, ``for certain'' has both senses
of ``certain'': ``particular'' and ``sure''.}
How should a
word like ``plant'' be categorized if it has uses both as a verb and
as a noun? How can a categorization be considered meaningful if the
infinitive marker ``to'' is not distinguished from the homophonous
preposition?

In a previous paper \cite{schohio},
we trained
a neural network to disambiguate
part-of-speech using context;
however,
no information about the word that is to be categorized was used. This
scheme fails for cases like ``The soldiers {\em rarely\/} come
home.'' vs. ``The soldiers {\em will\/} come home.''  where the context is
identical and information about the lexical item in question
(``rarely'' vs. ``will'') is needed in combination with context for
correct classification. In this paper, we will compare
two tagging algorithms, one based on classifying word types,
and one based on classifying words-plus-context.

\section{Tag induction}
We start by constructing representations of the syntactic behavior of
a word with respect to its left and right context.
Our working hypothesis is
that syntactic behavior is reflected in co-occurrence patterns.
Therefore, we will measure the similarity between two words with
respect to their syntactic behavior to, say, their left side by the
degree to which they share the same neighbors on the left. If the
counts of neighbors are assembled into a vector (with one dimension
for each neighbor), the cosine can be
employed to measure similarity.  It will assign a value close to 1.0
if two words share many neighbors, and 0.0 if they share none. We
refer to the vector of left neighbors of a word as its {\em left
context vector}, and to the vector of right neighbors as its {\em
right context vector}.
The unreduced context vectors in the experiment
described here have 250 entries, corresponding to the 250 most
frequent words in the Brown corpus.

This basic idea of measuring distributional similarity in terms of
shared neighbors
must be modified because of the
sparseness of the data. Consider
two infrequent adjectives that happen to modify different nouns in the
corpus. Their right similarity according to the cosine measure would be
zero. This is clearly undesirable. But even with high-frequency words,
the simple vector model can yield misleading similarity measurements. A
case in point is ``a'' vs.\ ``an''. These two articles do not share
any right neighbors since the former is only used before consonants
and the latter only before vowels. Yet intuitively, they are similar with
respect to their right syntactic context despite the lack of common
right neighbors.

Our solution to these problems is the application of a singular value
decomposition. We can represent the left vectors of all words in the
corpus as a matrix $C$ with $n$ rows, one for each word whose left
neighbors are to be represented, and $k$
columns, one for each of the possible neighbors.
SVD can be used to approximate the row and column vectors of $C$ in a
low-dimensional space. In more detail,
SVD decomposes  a matrix $C$, the matrix of left vectors
in our case, into
three matrices $T_{0}$, $S_{0}$, and $D_{0}$ such that:
\[C= T_{0} S_{0} D_{0}'\]
$S_{0}$ is a diagonal k-by-k matrix that contains the singular values of $C$
in descending order. The {\em i\/}th singular value can be interpreted
as indicating the strength of the {\em i\/}th principal component of
$C$.  $T_{0}$ and $D_{0}$ are orthonormal matrices
that approximate the rows and columns of $C$, respectively.  By
restricting the matrices $T_{0}$, $S_{0}$, and
$D_{0}$ to
their first $m<k$ columns (= principal components) one obtains the
matrices $T$, $S$, and $D$. Their product $\hat{C}$ is the best least
square
approximation of $C$ by a matrix of rank $m$: $\hat{C} = T S D' $.
We chose $m=50$ (reduction to a 50-dimensional space) for the SVD's
described in this paper.

SVD addresses the problems of generalization and sparseness because
broad and stable generalizations are represented on dimensions
with large values which will be retained in the dimensionality reduction.
In contrast, dimensions corresponding to small singular values
represent idiosyncrasies,  like the phonological constraint on the usage
of ``an'' vs.\ ``a'', and will be dropped.
We also gain efficiency since we can manipulate smaller vectors,
reduced to 50 dimensions.
We used SVDPACK to compute the singular
value decompositions described in this paper \cite{ber92}.

Table~\ref{nearest} shows the nearest neighbors of two words (ordered
according to closeness to the head word) after the dimensionality
reduction. Neighbors with highest similarity according to both left
and right context are listed. One can see clear differences between
the nearest neighbors in the two spaces. The right-context neighbors
of ``onto'' contain verbs because both prepositions and verbs govern
noun phrases to their right. The left-context neighborhood of ``onto''
reflects the fact that prepositional phrases are used in the same
position as adverbs like ``away'' and ``together'', thus making their
left context similar. For ``seemed'', left-context neighbors are words
that have similar types of noun phrases in subject position (mainly
auxiliaries).  The right-context neighbors all take ``to''-infinitives
as complements. An adjective like ``likely'' is very similar to
``seemed'' in this respect although its left context is quite
different from that of ``seemed''.  Similarly, the generalization that
prepositions and transitive verbs are very similar if not identical in
the way they govern noun phrases would be lost if ``left'' and
``right'' properties of words were lumped together in one
representation.  These examples demonstrate the importance of
representing generalizations about left and right context separately.

The left and right context vectors are the basis for four different
tag induction experiments, which are described in detail below:
\begin{itemize}
\item induction based on word type only
\item induction based on word type and context
\item induction based on word type and context, restricted to ``natural''
contexts
\item induction based on word type and context, using generalized left
and right context vectors
\end{itemize}

\subsection{Induction based on word type only}
The two context vectors of a word characterize the distribution of
neighboring words to its left and right.  The concatenation of left
and right context vector can therefore serve as a representation of a
word's distributional behavior \cite{fc92,schohio}. We formed such
concatenated vectors for all 47,025 words (surface forms) in the Brown corpus.
Here,
we use the raw 250-dimensional context vectors and apply the SVD to
the 47,025-by-500 matrix (47,025 words with two 250-dimensional
context vectors each). We obtained 47,025 50-dimensional reduced vectors from
the SVD and clustered them into 200 classes using the fast clustering
algorithm Buckshot \cite{cut92} (group average agglomeration applied
to a sample). This classification constitutes the baseline performance
for distributional part-of-speech tagging. All occurrences of a word
are assigned to one class. As pointed out above, such a procedure is
problematic for ambiguous words.

\subsection{Induction based on word type and context}
In order to exploit contextual information in the classification of a
token, we simply use context vectors of the two words occurring next
to the token.
An occurrence of word $w$ is represented by a concatenation of four
context vectors:
\begin{itemize}
\item The right context vector of the preceding word.
\item The left context vector of $w$.
\item The right context vector of $w$.
\item The left context vector of the following word.
\end{itemize}
The motivation is that a word's syntactic role depends both on
the syntactic properties of its neighbors and on
its own potential for entering into syntactic relationships with these
neighbors. The only properties of context that we consider
are the  right-context vector of the preceding
word and the left-context vector of the following word because they
seem to represent the contextual information most important for the
categorization of $w$. For example, for the disambiguation of ``work''
in ``her work seemed to be important'', only the fact that ``seemed''
expects noun phrases to its left is important, the right context
vector of ``seemed'' does not contribute to disambiguation.
That only the immediate neighbors are crucial for categorization is
clearly a simplification, but as the results presented below show it
seems to work surprisingly well.

\begin{table*}[tbp]
\small
\centering
\begin{tabular}{l|ll||l|ll}
tag & description & Penn Treebank tags &
 tag & description & Penn Treebank tags\\\hline

ADN & adnominal modifier & ADN$^{*}$ \$ &
POS & possessive marker & POS\\
CC & conjunction & CC &
PRP & pronoun & PRP\\
CD & cardinal & CD &
RB & adverbial & RB RP RBR RBS\\
DT &  determiner & DT PDT PRP\$ &
TO & infinitive marker & TO\\
IN & preposition & IN &
VB & infinitive & VB \\
ING & ``-ing'' forms & VBG &
VBD & inflected verb form & VBD VBZ VBP \\
MD & modal & MD &
VBN & predicative & VBN PRD$^{*}$ \\
N & nominal & NNP(S) NN(S) &
WDT & wh-word & WP(\$) WRB WDT
\end{tabular}
\caption{\label{tagset} Evaluation tag set.
Structural tags derived from parse trees are marked with $*$.
}
\end{table*}

Again, an SVD is applied to address the problems of sparseness and
generalization. We randomly selected 20,000 word triplets from the corpus
and formed concatenations of four context vectors as described above.
The singular value decomposition of the resulting 20,000-by-1,000
matrix defines a mapping from the 1,000-dimensional space of
concatenated context vectors to a 50-dimensional reduced space.  Our
tag set was then induced by clustering the reduced vectors of the
20,000 selected occurrences into 200 classes. Each
of the 200 tags is defined by the centroid of the corresponding class
(the sum of its members).  Distributional tagging of an occurrence of
a word $w$ proceeds then by retrieving the four relevant context
vectors (right context vector of previous word, left context vector of
following
word, both context vectors of $w$) concatenating them to one
1000-component vector, mapping this vector to 50 dimensions,
computing the correlations with the 200 cluster centroids and,
finally, assigning the occurrence to the closest cluster. This
procedure was applied to all tokens of the Brown corpus.

We will see below that this method of distributional tagging, although
partially successful, fails for many tokens whose neighbors are
punctuation marks. The context vectors of punctuation marks contribute
little information about syntactic categorization since there are no
grammatical dependencies between words and punctuation marks, in
contrast to strong dependencies between neighboring words.

For this reason, a second induction on the basis of word type and
context was performed, but only for those tokens with informative
contexts. Tokens next to punctuation marks and tokens with rare words
as neighbors were not included. Contexts with rare words (less than
ten occurrences) were also excluded for similar reasons: If a word
only occurs nine or fewer times its left and right context vectors
capture little information for syntactic categorization. In the
experiment,
20,000
natural contexts were randomly selected, processed by the SVD and
clustered into 200 classes. The classification was then applied to all
natural contexts of the Brown corpus.

\subsection{Generalized context vectors}
The context vectors used so far only capture information about
distributional interactions with the 250 most frequent words.
Intuitively, it should be possible to gain accuracy in tag induction
by using information from more words. One way to do this is to let the
right context vector record which {\em classes of left context
vectors} occur to the right of a word.  The rationale is that words
with similar left context characterize words to their right in a
similar way. For example, ``seemed'' and ``would'' have similar left
contexts, and they characterize the right contexts of ``he'' and ``the
firefighter'' as potentially containing an inflected verb form. Rather
than having separate entries in its right context vector for
``seemed'', ``would'', and ``likes'', a word like ``he'' can now be
characterized by a generalized entry for ``inflected verb form occurs
frequently to my right''.

This proposal was implemented by applying a singular value
decomposition to the 47025-by-250 matrix of left context vectors and
clustering the resulting context vectors into 250 classes. A
generalized right context vector $v$ for word $w$ was then formed by
counting how often words from these 250 classes occurred to the right of $w$.
Entry $v_{i}$ counts the number of times that a word from class $i$
occurs to the right of $w$ in the corpus (as opposed to the number of
times that the word with frequency rank $i$ occurs to the right of
$w$).  Generalized left context vectors were derived by an analogous
procedure using word-based right context vectors.  Note that the
information about left and right is kept separate in this computation.
This differs from previous approaches \cite{fc92,schohio} in which
left and right context vectors of a word are always used in one
concatenated vector. There are arguably fewer different types of right
syntactic contexts than types of syntactic categories. For example,
transitive verbs and prepositions belong to different syntactic
categories, but their right contexts are virtually identical in that
they require a noun phrase. This generalization could not be exploited
if left and right context were not treated separately.

Another argument for the two-step derivation is that many
words don't have any of the 250 most frequent words as their left or
right neighbor. Hence, their vector would be zero in the word-based
scheme. The class-based scheme makes it more likely that
meaningful
representations are formed for all words in the vocabulary.

The generalized context vectors were input to the
tag induction procedure described above for word-based context
vectors: 20,000 word triplets were selected from the corpus, encoded as
1,000-dimensional vectors (consisting of four generalized context
vectors), decomposed by a singular value decomposition and clustered
into 200 classes. The resulting classification was applied to all
tokens in the Brown corpus.

\section{Results}
The results of the four experiments were evaluated by forming 16
classes of tags from the Penn Treebank as shown in Table~\ref{tagset}.
Preliminary experiments showed that distributional
methods distinguish adnominal and predicative uses of
adjectives (e.g.\ ``the black cat'' vs.\ ``the cat is black'').
Therefore the tag ``ADN'' was introduced for
uses of adjectives, nouns, and participles as adnominal modifiers.
The tag ``PRD'' stands for predicative uses of adjectives.
The Penn Treebank parses of the Brown corpus were used to determine
whether a token functions as an adnominal modifier.
Punctuation marks, special symbols, interjections, foreign words and
tags with fewer than 100 instances were excluded from the
evaluation.

\begin{table*}[tbp]
\small
\centering
\begin{tabular}{l||r|r|r|r|r|r|l}
tag &frequency &\# classes &correct&incorrect &precision &recall &
F\\\hline\hline

ADN  &108586 &34 & 38282 & 19528 &0.66 & 0.35 & 0.46 \\
CC  &36808 &0 & 0 & 0 &0.00 & 0.00 & 0.00 \\
CD  &15085 &4 & 3376 & 1431 &0.70 & 0.22 & 0.34 \\
DT  &129626 &2 & 125540 & 31783 &0.80 & 0.97 & 0.87 \\
IN  &132079 &3 & 118726 & 75829 &0.61 & 0.90 & 0.73 \\
ING  &14753 &5 & 2111 & 1016 &0.68 & 0.14 & 0.24 \\
MD  &13498 &2 & 13383 & 13016 &0.51 & 0.99 & 0.67 \\
N  &231434 &98 & 193838 & 79652 &0.71 & 0.84 & 0.77 \\
POS  &5086 &1 & 4641 & 1213 &0.79 & 0.91 & 0.85 \\
PRP  &47686 &3 & 43839 & 21723 &0.67 & 0.92 & 0.77 \\
RB  &54525 &7 & 35364 & 56505 &0.38 & 0.65 & 0.48 \\
TO  &25196 &0 & 0 & 0 &0.00 & 0.00 & 0.00 \\
VB  &35342 &8 & 29138 & 17945 &0.62 & 0.82 & 0.71 \\
VBD  &80058 &12 & 36653 & 3855 &0.90 & 0.46 & 0.61 \\
VBN  &41146 &21 & 7773 & 8841 &0.47 & 0.19 & 0.27 \\
WDT  &14093 &0 & 0 & 0 &0.00 & 0.00 & 0.00 \\
\hline
avg. &&&&& 0.53 & 0.52 & 0.49

\end{tabular}
\caption{\label{wordtype} Precision and recall for induction based on
word type.}
\end{table*}

Tables~\ref{wordtype} and \ref{context} present results for word
type-based induction and induction based on word type and context.
For each tag $t$, the table lists the frequency
of $t$ in the corpus (``frequency'')\footnote{The small difference in
overall frequency in the tables is due to the fact that some
word-based context vectors consist entirely of zeros. There were about
a hundred word triplets whose four context vectors did not have
non-zero entries and could not be assigned a cluster.}, the number of induced
tags
$i_{0}, i_{1},
\ldots,i_{l}$, that were assigned to it (``\# classes''); the number
of times an occurrence of $t$ was correctly labeled as belonging to
one of $i_{0},i_{1},\ldots,i_{l}$ (``correct''); the number of times
that a token of a different tag $t^{\prime}$ was miscategorized as being an
instance of $i_{0},i_{1},\ldots,i_{l}$ (``incorrect''); and precision
and recall of the categorization of $t$. Precision is the number of
correct tokens divided by the sum of correct and incorrect tokens.
Recall is the number of correct tokens divided by the total number
of tokens of $t$
(in the first column). The last column gives van Rijsbergen's
F measure which computes an aggregate score from precision and
recall: \cite{rij79}
$ F = \frac{1}{\alpha \frac{1}{P} + (1-\alpha) \frac{1}{R}} $.
We chose $\alpha = 0.5$ to give equal weight to precision and recall.

It is clear from the tables that incorporating context improves
performance considerably. The F score increases for all tags except
CD, with an average improvement of more than 0.20. The tag CD is
probably
better thought of as describing a word class. There is a wide range of
heterogeneous
syntactic functions of cardinals in particular contexts:
quantificational and adnominal uses, bare NP's (``is one of''), dates
and ages (``Jan 1'', ``gave his age as 25''), and enumerations.
In this light, it is not surprising that the word-type method does
better on cardinals.

\begin{table*}[tbp]
\small
\centering
\begin{tabular}{l||r|r|r|r|r|r|l}
tag &frequency &\# classes &correct&incorrect &precision &recall &
F\\\hline\hline

ADN  &108532 &42 & 87128 & 24743 &0.78 & 0.80 & 0.79 \\
CC  &36808 &2 & 28671 & 1501 &0.95 & 0.78 & 0.86 \\
CD  &15084 &1 & 747 & 809 &0.48 & 0.05 & 0.09 \\
DT  &129626 &6 & 119534 & 6178 &0.95 & 0.92 & 0.94 \\
IN  &132079 &11 & 125554 & 25316 &0.83 & 0.95 & 0.89 \\
ING  &14753 &4 & 3096 & 4876 &0.39 & 0.21 & 0.27 \\
MD  &13498 &2 & 12983 & 936 &0.93 & 0.96 & 0.95 \\
N  &231424 &68 & 207822 & 51695 &0.80 & 0.90 & 0.85 \\
POS  &5086 &2 & 4623 & 533 &0.90 & 0.91 & 0.90 \\
PRP  &47686 &7 & 44946 & 12759 &0.78 & 0.94 & 0.85 \\
RB  &54524 &16 & 31184 & 17403 &0.64 & 0.57 & 0.60 \\
TO  &25196 &1 & 23291 & 61 &1.00 & 0.92 & 0.96 \\
VB  &35342 &8 & 29392 & 6152 &0.83 & 0.83 & 0.83 \\
VBD  &80058 &17 & 64150 & 8663 &0.88 & 0.80 & 0.84 \\
VBN  &41145 &11 & 25578 & 11972 &0.68 & 0.62 & 0.65 \\
WDT  &14093 &2 & 1621 & 1017 &0.61 & 0.12 & 0.19 \\
\hline
avg. &&&&& 0.78 & 0.71 & 0.72

\end{tabular}
\caption{\label{context} Precision and recall for induction based on
word type and context.}
\end{table*}

Table~\ref{general} shows that performance for generalized context
vectors is better than for word-based context vectors (0.74 vs.\
0.72). However, since the number of tags with better and worse
performance is about the same (7 and 5), one cannot conclude with
certainty that generalized context vectors induce tags of higher
quality. Apparently, the 250 most frequent words capture most of the
relevant distributional information so that the additional information
from less frequent words available from generalized vectors only has a
small effect.

\begin{table*}[tbp]
\small
\centering
\begin{tabular}{l||r|r|r|r|r|r|l}
tag &frequency &\# classes &correct&incorrect &precision &recall &
F\\\hline\hline

ADN  &108586 &50 & 91893 & 26790 &0.77 & 0.85 & 0.81 \\
CC  &36808 &4 & 34127 & 6430 &0.84 & 0.93 & 0.88 \\
CD  &15085 &3 & 3707 & 1530 &0.71 & 0.25 & 0.36 \\
DT  &129626 &10 & 120968 & 5780 &0.95 & 0.93 & 0.94 \\
IN  &132079 &8 & 123516 & 22070 &0.85 & 0.94 & 0.89 \\
ING  &14753 &2 & 3798 & 7161 &0.35 & 0.26 & 0.30 \\
MD  &13498 &3 & 13175 & 1059 &0.93 & 0.98 & 0.95 \\
N  &231434 &70 & 201890 & 33206 &0.86 & 0.87 & 0.87 \\
POS  &5086 &2 & 4932 & 1636 &0.75 & 0.97 & 0.85 \\
PRP  &47686 &5 & 37535 & 9221 &0.80 & 0.79 & 0.79 \\
RB  &54524 &9 & 29892 & 18398 &0.62 & 0.55 & 0.58 \\
TO  &25196 &1 & 25181 & 27 &1.00 & 1.00 & 1.00 \\
VB  &35342 &7 & 28879 & 6560 &0.81 & 0.82 & 0.82 \\
VBD  &80058 &15 & 66457 & 12079 &0.85 & 0.83 & 0.84 \\
VBN  &41145 &10 & 26960 & 17356 &0.61 & 0.66 & 0.63 \\
WDT  &14093 &1 & 2223 & 563 &0.80 & 0.16 & 0.26 \\
\hline
avg. &&&&& 0.78 & 0.73 & 0.74

\end{tabular}
\caption{\label{general} Precision and recall for induction based on
generalized context vectors.}
\end{table*}

\begin{table*}[tbp]
\small
\centering
\begin{tabular}{l||r|r|r|r|r|r|l}
tag &frequency &\# classes &correct&incorrect &precision &recall &
F\\\hline\hline

ADN  &63771 &36 & 54398 & 12203 &0.82 & 0.85 & 0.83 \\
CC  &16148 &4 & 15657 & 1798 &0.90 & 0.97 & 0.93 \\
CD  &7011 &1 & 1857 & 918 &0.67 & 0.26 & 0.38 \\
DT  &87914 &9 & 82206 & 2664 &0.97 & 0.94 & 0.95 \\
IN  &91950 &9 & 86793 & 6842 &0.93 & 0.94 & 0.94 \\
ING  &7268 &2 & 1243 & 1412 &0.47 & 0.17 & 0.25 \\
MD  &11244 &3 & 10363 & 476 &0.96 & 0.92 & 0.94 \\
N  &111368 &49 & 100105 & 14452 &0.87 & 0.90 & 0.89 \\
POS  &3202 &1 & 2912 & 255 &0.92 & 0.91 & 0.91 \\
PRP  &23946 &7 & 22877 & 4062 &0.85 & 0.96 & 0.90 \\
RB  &32331 &16 & 21037 & 9922 &0.68 & 0.65 & 0.66 \\
TO  &19859 &2 & 19537 & 53 &1.00 & 0.98 & 0.99 \\
VB  &26714 &11 & 24036 & 4119 &0.85 & 0.90 & 0.88 \\
VBD  &56540 &33 & 51016 & 8488 &0.86 & 0.90 & 0.88 \\
VBN  &24804 &14 & 18889 & 7448 &0.72 & 0.76 & 0.74 \\
WDT  &8329 &3 & 3691 & 670 &0.85 & 0.44 & 0.58 \\
\hline
avg. &&&&& 0.83 & 0.78 & 0.79

\end{tabular}
\caption{\label{natural} Precision and recall for induction for
natural contexts.}
\end{table*}

Table~\ref{natural} looks at results for ``natural'' contexts, i.e.\
those not containing punctuation marks and rare words. Performance is
consistently better than for the evaluation on all contexts,
indicating that the low quality of the distributional information
about punctuation marks and rare words is a difficulty for successful
tag induction.

Even for ``natural'' contexts,
performance varies considerably. It is fairly good for prepositions,
determiners, pronouns, conjunctions, the infinitive marker, modals,
and the possessive marker. Tag induction fails for cardinals (for the reasons
mentioned above) and for ``-ing'' forms. Present participles and gerunds
are difficult because they exhibit both verbal and nominal properties
and occur in a wide variety of different contexts whereas other parts
of speech have a few typical and frequent contexts.

It may seem worrying that some of the tags are assigned a high number
of clusters (e.g., 49 for N, 36 for ADN).
A closer look
reveals that many clusters embody finer distinctions. Some examples:
Nouns in
 cluster 0 are heads of larger noun phrases, whereas the nouns
in cluster 1 are full-fledged NPs.  The members of classes 29
and 111 function as subjects. Class 49 consists of proper nouns.
However, there are many pairs or triples of clusters that
should be collapsed into one on linguistic grounds. They were
separated on distributional criteria that don't have linguistic
correlates.

An analysis of the divergence between our classification and the
manually assigned tags revealed three main sources of errors:
rare words and rare syntactic phenomena,
 indistinguishable distribution,
and non-local dependencies.

Rare words are difficult because of lack of distributional evidence.
For example, ``ties'' is used as a verb only 2 times (out of 15
occurrences in the corpus). Both occurrences are miscategorized, since
its context vectors do not provide enough evidence for the verbal use.
Rare
syntactic constructions pose a related problem: There are not enough
instances to justify the creation of a separate cluster. For example,
verbs taking bare infinitives were classified as adverbs since this
is too rare a phenomenon to provide strong distributional evidence
(``we do not DARE speak of'', ``legislation could HELP remove'').

The case of the tags ``VBN'' and ``PRD'' (past participles and
predicative adjectives)
demonstrates the difficulties of word classes
with indistinguishable distributions. There are hardly any
distributional clues for distinguishing ``VBN'' and ``PRD'' since both
are mainly used as complements of ``to be''.\footnote{Because of
phrases like ``I had sweet potatoes'',
forms of ``have'' cannot serve as a reliable discriminator either.}
A common tag class was created for
``VBN'' and ``PRD'' to show that they are reasonably well
distinguished from other parts of speech, even if not from each other.
Semantic understanding is necessary to distinguish
between the states described by phrases of the form ``{\em to be\/}
adjective''  and the processes described by
phrases of the form ``{\em to be\/} past participle''.

Finally, the method fails if there are no local dependencies that
could be used for categorization and only non-local dependencies are
informative. For example,
the adverb in ``Mc*N. Hester,
CURRENTLY Dean of \ldots'' and the conjunction in
``to
add that, IF United States policies \ldots'' have similar immediate
neighbors (comma, NP).
The decision to consider only immediate neighbors is
responsible for this type of error since taking a wider context into account
would disambiguate the parts of speech in question.

\section{Future Work}
There are three avenues of future research we are interested in
pursuing. First, we are planning to apply the algorithm to an as yet
untagged language. Languages with a rich morphology may be
more difficult than English since with fewer tokens per type, there
is less data on which to base a categorization decision.

Secondly, the error analysis suggests that considering non-local
dependencies would improve results. Categories that can be induced
well (those characterized by local dependencies) could be input into
procedures that learn phrase structure (e.g. \cite{bm92,fin93}).
These phrase constraints could then be incorporated into the
distributional tagger to characterize non-local dependencies.

Finally, our procedure induces a ``hard'' part-of-speech
classification of occurrences in context, i.e., each occurrence is
assigned to only one category. It is by no means generally accepted
that such a classification is linguistically adequate. There is both
synchronic \cite{ros72} and diachronic \cite{tab94} evidence
suggesting that words and their uses can inherit properties from
several prototypical syntactic categories. For example, ``fun'' in
``It's a fun thing to do.'' has properties of both a noun and an
adjective (superlative ``funnest'' possible). We are planning to
explore ``soft'' classification algorithms that can account for these
phenomena.

\section{Conclusion}
In this paper, we have attempted to construct an algorithm for fully
automatic distributional tagging, using unannotated corpora as the
sole source of information. The main innovation is that the algorithm
is able to deal with part-of-speech ambiguity, a pervasive phenomenon
in natural language that was unaccounted for in previous work on learning
categories from corpora.  The method was systematically evaluated on
the Brown corpus.  Even if no automatic procedure can rival the
accuracy of human tagging, we hope that the algorithm will facilitate
the initial tagging of texts in new languages and sublanguages.

\section{Acknowledgments}
I am grateful for helpful comments
to Steve Finch, Jan Pedersen and
two anonymous reviewers (from ACL and EACL).
I'm also indebted to Michael Berry for SVDPACK and to the Penn
Treebank Project
for the parsed Brown corpus.

\end{document}